%% file: main.tex
\RequirePackage{silence}
\documentclass[runningheads]{llncs}

\usepackage{xcolor}
\usepackage{graphicx}
\graphicspath{{/}{fig/}}
\usepackage{cite}
\usepackage{pgfplots} 
\usepackage{array}
\usepackage{textcomp}
\usepackage{xcolor}
\usepackage{multirow}
\usepackage{booktabs}
\usepackage{float}
\usepackage{enumerate}
\usepackage{subcaption}
\usepackage[vlined, ruled, shortend]{algorithm2e} 
\usepackage{mathtools}
\usepackage{gensymb}

\usetikzlibrary{spy}

\newlength\figureheight
\newlength\figurewidth
\SetAlCapNameFnt{\footnotesize}
\SetAlCapFnt{\footnotesize}

\usepackage{tikz}

\begin{document}
\title{Evaluating the Performance of Multi-Scan Integration for UAV LiDAR-based Tracking}
%
%
%
\author{
    Iacopo Catalano\inst{1} \and
    Jorge Peña Queralta\inst{1} \and
    Tomi Westerlund\inst{1}
}%
\authorrunning{Iacopo Catalano et al.}
\titlerunning{Multi-Scan Integration for UAV LiDAR-based Tracking}
\institute{
        Turku Intelligent Embedded and Robotic Systems \\
        University of Turku, Finland\\ 
        \email{\{imcata, jopequ, tovewe\}@utu.fi} \\
        \url{https://tiers.utu.fi}
    }
\maketitle
%
%
%
\input{sections/00_Abstract}

\input{sections/01_Introduction}

\input{sections/02_RelatedWork}
\input{sections/03_Methodology}
\input{sections/04_Experiments}
\input{sections/05_Conclusion}

\section*{Acknowledgment}

This research work is supported by the Academy of Finland's AeroPolis project (Grant No. 348480).


\bibliographystyle{unsrt}
\bibliography{bibliography}

\end{document}

%% file: sections/00_Abstract.tex
\begin{abstract}

    Drones have become essential tools in a wide range of industries, including agriculture, surveying, and transportation. However, tracking unmanned aerial vehicles (UAVs) in challenging environments, such cluttered or GNSS-denied environments, remains a critical issue. Additionally, UAVs are being deployed as part of multi-robot systems, where tracking their position can be essential for relative state estimation. In this paper, we evaluate the performance of a multi-scan integration method for tracking UAVs in GNSS-denied environments using a solid-state LiDAR and a Kalman Filter (KF). We evaluate the algorithm's ability to track a UAV in a large open area at various distances and speeds. Our quantitative analysis shows that while "tracking by detection" using a Constant Velocity model is the only method that consistently tracks the target, integrating multiple scan frequencies using a KF achieves lower position errors and represents a viable option for tracking UAVs in similar scenarios.

\keywords{
    UAV         \and 
    Tracking         \and 
    LiDAR         \and 
    Multi-Scan Integration         \and 
}

\end{abstract}

%% file: sections/01_Introduction.tex
\section{Introduction}
\label{sec:intro}

Unmanned Aerial Vehicles (UAVs) have become increasingly popular in various domains, owing to their mobility and versatility as mobile sensing platforms~\cite{tsouros2019review, wang2019surveying}. They are commonly used for aerial photography, mapping, surveying, delivery, search and rescue, among others~\cite{queralta2020collaborative, osco2021review}. However, tracking UAVs in challenging environments, such as Global Navigation Satellite System (GNSS)-denied areas or cluttered environments, remains a difficult task~\cite{li2020autotrack, jiang2021anti, hasier2023uav}. GNSS signals may not be available in certain areas, such as indoor environments or urban canyons, which limits the accuracy and reliability of UAV tracking. Moreover, they can be susceptible to jamming or spoofing, which can lead to severe consequences such as loss of control or collisions with other objects.

UAVs are also increasingly part of real-world multi-robot systems, where tracking between robots is often part of relative or global state estimation approaches~\cite{queralta2022vio, bai2020cooperative}. Indeed, the deployment of multi-robot systems in GNSS-denied environments has been highlighted as an important and recent development~\cite{queralta2020collaborative}. An illustrative case in point is the DARPA Subterranean challenge, which has attracted considerable attention~\cite{rouvcek2019darpa, petrlik2020robust}. According to the reports from the participating teams, the successful localization and collaborative sensing were identified as some of the key challenges faced during the competition. Notably, UAVs were dynamically dispatched from unmanned ground vehicles (UGVs) during the challenge, which further underscores the complex nature of the deployment.

In recent years, researchers have shown a growing interest in tracking and detecting UAVs due to two main reasons. The first is the pressing need to identify and detect foreign objects or drones in areas with controlled airspace, including airports~\cite{guvenc2018detection, hengy2017multimodal}. The second reason is the potential for optimizing the utilization of UAVs as flexible mobile sensing platforms~\cite{queralta2020autosos}.

\begin{figure}[]
    \centering
    \includegraphics[width=\textwidth]{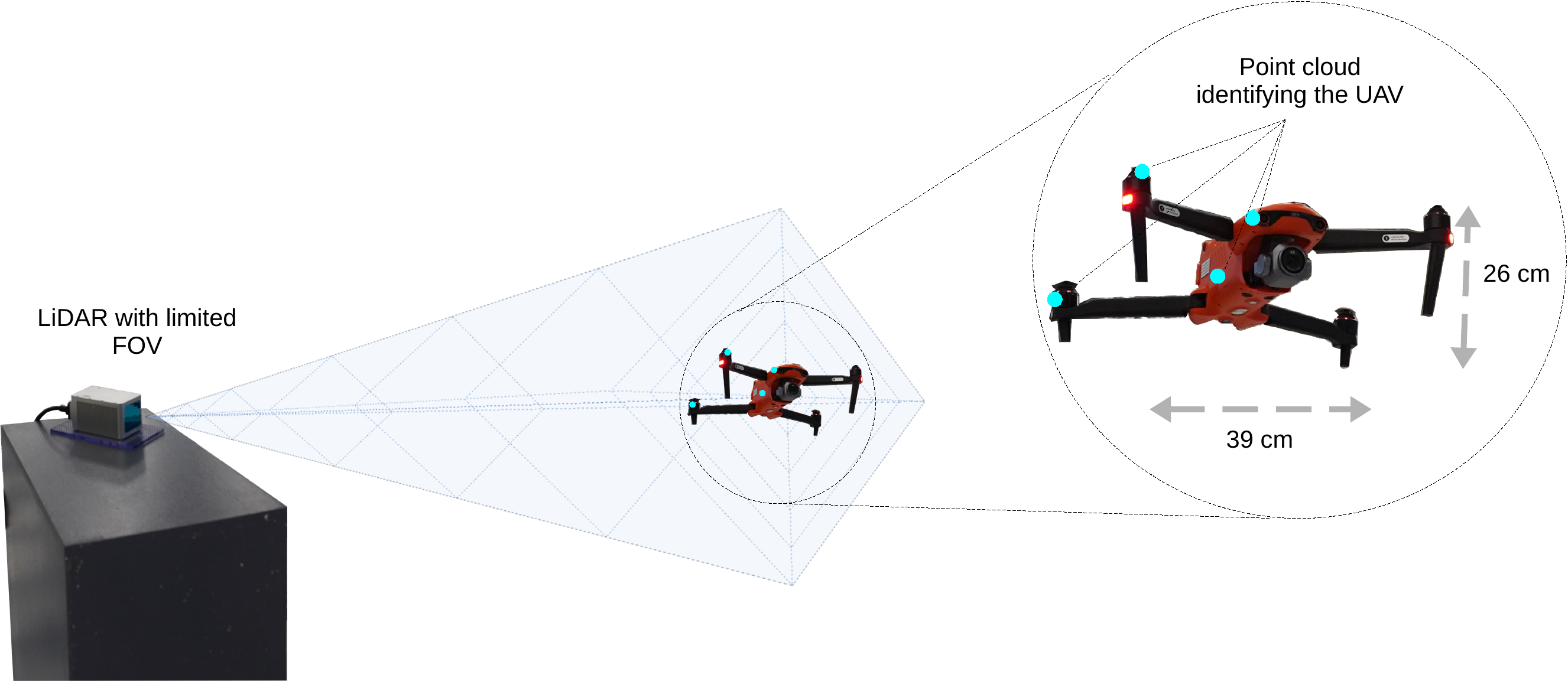}
    \caption{UAV tracking using a limited-FoV solid-state LiDAR.}
    \label{fig:lidar_mav_tracking}
\end{figure}

While significant progress has been made in UAV tracking using GNSS and other sensors, existing methods are not without limitations. Many methods may rely on expensive hardware or require high levels of computational power, which limits their practicality and scalability.

Solid-state LiDARs are a promising technology in long-range scanning that produce high-density point clouds, making them ideal for tracking objects in three-dimensional space, such as UAVs~\cite{li2020towards, qingqing2020towards}. Their main limitation is the restricted field of view (FoV). This is illustrated in Fig.~\ref{fig:lidar_mav_tracking}. However, solutions to this problem have been proposed, such as utilizing multiple LiDARs or adjusting the position and orientation of the robot base where the LiDAR is installed, in order to compensate for this limitation~\cite{van2021solid, qingqing2022multi}. The frame or scan frequency concept differs significantly in these LiDARs from the more standard spinning 3D LiDARs. In the latter, a frame can be naturally generated aggregating laser measurements from a single revolution. Solid-state LiDARs, with non-repetitive scan patterns, can output point clouds at adjustable frequencies with varying FoV coverage. This characteristic creates new opportunities for developing LiDAR perception methods that allow for the adaptive adjustment of the frame integration time in order to improve object sensing capabilities.

In this article, we present an evaluation of the multi-scan integration method for tracking UAVs in challenging environments using a solid-state LiDAR (see Fig.~\ref{fig:lidar_mav_tracking}). We build upon our previous work~\cite{qingqing2021adaptive} where we presented a qualitative analysis without real tracking, by proposing both a linear and an Extended Kalman Filter for tracking UAVs with a Constant Velocity motion model. Unlike the previous work, where we used "tracking by detection" to estimate the target's future position, we provide a more sophisticated approach that enables real tracking of UAVs. Additionally, in this work we are able to quantify the tracking performance with ground truth data generated by a motion capture (MOCAP) system during the experiments.

To evaluate our method, we integrate different numbers of scans, ranging from 2 scans to 50 scans, to track the UAV's position accurately. By setting the sensor scan rate to its maximum, 100\,Hz, we are able to effectively scan the environment at frequencies ranging from 50\,Hz (2 scans) to 2\,Hz (50 scans), simply by accumulating scans before processing the data. Our results showcase the limitations of the multi-scan integration method in dynamic environments and multi-UAV systems, and motivate further research to address the limitations and enhance the method's performance.

In the following sections, we will begin by providing a brief review of related work on UAV tracking and LiDAR-based sensing. Then, we will delve into the proposed method in detail and present experimental results that demonstrate the accuracy and robustness of the method in tracking UAVs. Finally, we will conclude by highlighting potential avenues for future research.

%% file: sections/02_RelatedWork.tex
\section{Related work}
\label{sec:related}

LiDAR systems are often employed for detecting and tracking objects, including UAVs. However, tracking UAVs with LiDAR can be challenging due to their small size, varied shapes and materials, high speed, and unpredictable movements.

In order to overcome these challenges, researchers have explored various methods to overcome the limitations of 3D LiDAR technology and improve the detection and tracking of UAVs. One approach involves conducting a probabilistic analysis of detections, as described in~\cite{dogru2022dronedetection}, which allows for achieving detection using fewer LiDAR beams while continuously tracking only a small number of hits. Moreover, increasing the field of view and improving the coverage ratio have been identified as effective means. A different strategy involves combining a segmentation approach and a simple object model while leveraging temporal information, as demonstrated in~\cite{razlaw2019DetectionAT}. This approach has been shown to reduce the parametrization effort and generalize well in different settings. Overall, the use of LiDAR technology offers various methods for improving the detection and tracking of UAVs, and researchers are continually exploring new techniques to overcome the unique challenges posed by these small, fast-moving, and unpredictable objects.

In the context of deploying UAVs from a ground robot, one critical factor to consider is the relative localization between different devices. In order to address this, Li et al.~\cite{qingqing2021adaptive} proposed a novel approach for tracking UAVs using LiDAR point clouds. This approach takes into account the UAV's speed and distance from the sensor to dynamically adjust the LiDAR's frame integration time. This adjustment affects the density and size of the point cloud that needs to be processed.

Additionally, Sier et al~\cite{hasier2023uav} adopt the LiDAR-as-a-camera concept fusing images and point cloud data generated by a single LiDAR sensor to track UAVs without a priori knowledge. Employing a custom YOLOv5 model trained on panoramic images, they are capable of bringing computer vision capabilities on top of the LiDAR itself.

Another technique, departing from the typical sequence of track-after-detect, is to leverage motion information by searching for minor 3D details in the 360$^{\circ}$ LiDAR scans of the scene and analyzing the trajectory of the tracked object to classify UAVs and non-UAV objects by identifying typical movement patterns~\cite{hammer2018potentiallidardetection, hammer2018lidarsmalluavs}.

%% file: sections/03_Methodology.tex
\section{Methodology}

\begin{figure}
    \centering
        \includegraphics[width=\textwidth]{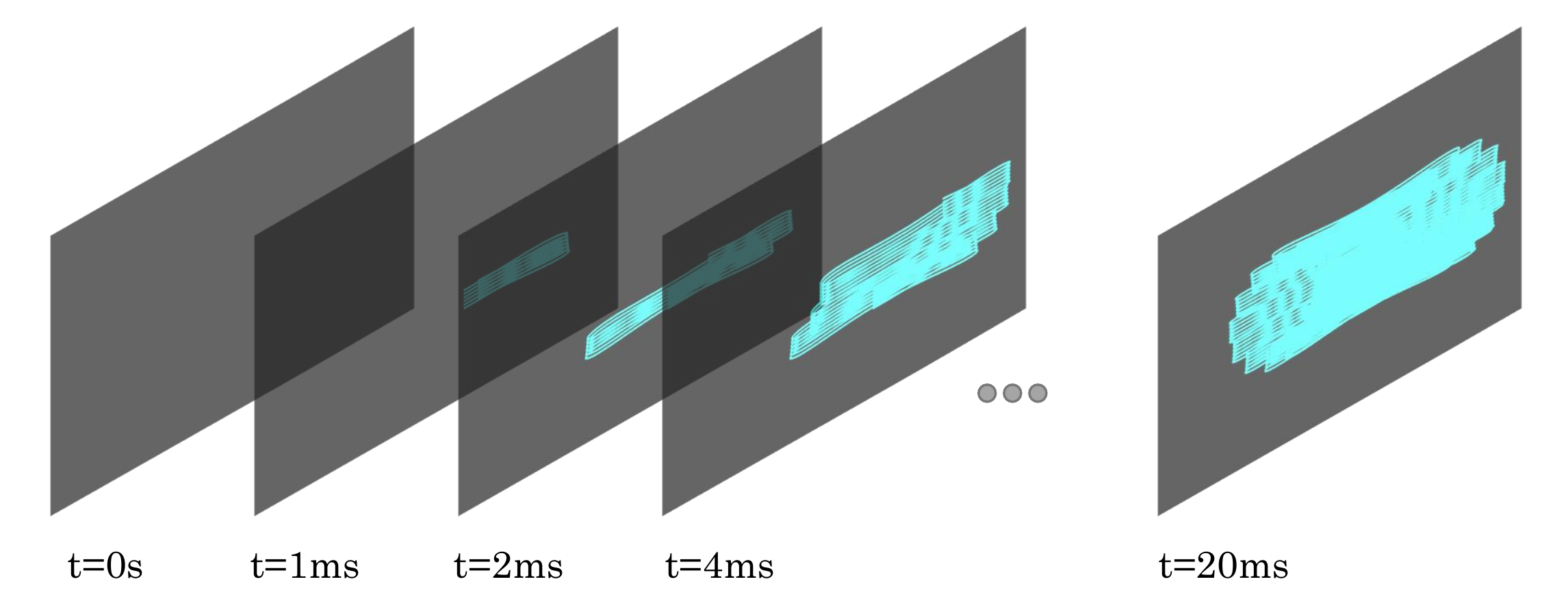}
        \caption{FoV coverage of the solid-state LiDAR used in this work with different point cloud integration times.}
        \label{fig:lidar_fov}
\end{figure}

The majority of 3D laser scanners currently available are multi-channel, rotating LiDARs. While high-end devices with 64 or 128 vertical channels can provide excellent angular resolution in both horizontal and vertical dimensions, they are not the most commonly used. Additionally, the repetitive scanning pattern of these devices has been beneficial from a geometric perspective in terms of data processing. However, it limits the FoV coverage and exposure time if the sensor position is fixed. Alternatively, solid state LiDARs with non-repetitive scan patterns can generate dense point clouds with adjustable frequencies and varying FoV coverage as illustrated in Fig.~\ref{fig:lidar_fov}, providing opportunities for new perception methods that make use of different multi-scan integration time ranges to better detect objects.

In the following formulation we will use discrete steps represented by $k$ owing to the discrete nature of the set of consecutive point clouds. Let $\mathcal{P}^{}_k(I_{r}^k) = \{ \mathbf{p}_{1}^{k}, \mathbf{p}_{2}^{k}, \dots, \mathbf{p}_{n_k}^{k} \}$ be the set of $n_k$ points in the point cloud generated by the LiDAR sensor at time step $k$ using an integration time $I_{r}^k$, where $r$ is the range of the interval. Each point $\mathbf{p}_{i}^{k}$ has a position and velocity vector $\mathbf{x}_{i}^{k} = [x_{i}^{k}, y_{i}^{k}, z_{i}^{k}]^{\top}$ and $\dot{\mathbf{x}}_{i}^{k} = [\dot{x}_{i}^{k}, \dot{y}_{i}^{k}, \dot{z}_{i}^{k}]^{\top}$ , respectively. We also denote by $\textbf{s}^k_{UAV}$=\{$\textbf{x}^{k}_{UAV}$,$\dot{\textbf{x}}^{k}_{UAV}$\} the position and speed of the UAV. The objective of the tracking algorithm is to identify the subset of points in $\mathcal{P}^{}_k(I_{r}^k)$ that corresponds to the UAV, denoted $\mathcal{P}^{k}_{UAV}$, in order to estimate its position and velocity.

To initiate the tracking process, we assume that the initial position of the UAV ($\textbf{x}^0_{UAV}$) is known. The point cloud $\mathcal{P}^{}_k(I_{r}^k)$ is integrated by accumulating the number of scans defined by $I_{r}^k$. We then employ a nearest-neighbor search algorithm to identify the points in the point cloud that are closest to the predicted position of the UAV, based on its initial position. We leverage a priori information about the dimensions of the tracked object to improve the reliability and accuracy of the tracking results: the nearest-neighbor search is constrained to a search radius $r$ around the initial position as represented in Equation~\ref{eq:nn_search}.

\begin{equation}\label{eq:nn_search}
    \mathcal{P}^k_{UAV} = {\textbf{x} \in \mathcal{P}^{}_k(I^k_r) : ||\textbf{x} - \textbf{x}^{}_{UAV}||_2 \leq R}
\end{equation}

This allows us to constrain the nearest-neighbor search to a smaller volume around the estimated position, leading to faster and more accurate search results.

Next, we average the extracted points to estimate a new position $\textbf{x}^k_{UAV}$ for the UAV, which serves as the measurement in the Kalman Filter's measurement update step as expressed in Equation~\ref{eq:points_avg}.

\begin{equation}\label{eq:points_avg}
    \hat{\textbf{x}}^{k}_{UAV} = \frac{1}{|\mathcal{P}^k_{UAV}|} \sum_{\textbf{x} \in \mathcal{P}^k_{UAV}} \textbf{x}
\end{equation}

For the prediction step of the Kalman Filter, we adopt a Constant Velocity (CV) motion model to obtain a new estimate $\hat{\textbf{x}}_{k|k-1}$ of the UAV's position and velocity at time $k$ as per Eq~\ref{eq:kf_prediction}:

\begin{equation}\label{eq:kf_prediction}
    \begin{aligned}
    \hat{\textbf{x}}_{k|k-1} &= \textbf{x}_{k-1} + \hat{\dot{\textbf{x}}}_{k-1} \Delta t \\
    \hat{\dot{\textbf{x}}}_{k|k-1} &= \hat{\dot{\textbf{x}}}_{k-1}
    \end{aligned}
\end{equation}

where $\Delta t$ is the time step between consecutive measurements, and $\hat{\textbf{x}}_{k|k-1}$ and $\hat{\dot{\textbf{x}}}_{k|k-1}$ are the predicted position and velocity, respectively.

We then repeat the nearest-neighbor search around the new predicted position to update the measurement in the next iteration of the Kalman Filter. This process is outlined in Algorithm~\ref{alg:tracking}.

\input{algs/tracking}

%% file: algs/tracking.tex
\begin{algorithm}[t]
    \footnotesize
	\caption{\footnotesize UAV tracking with scan integration}
	\label{alg:tracking}
	\KwIn{\\ 
	    \begin{tabular}{ll}
            Integration Rate:                 & $I$ \\[+0.4em]
	        3D LiDAR point cloud:             & $\mathcal{P}_{k}(I)$ \\[+0.4em]
	        Last known UAV state:             & $ (\textbf{x}^{k-1}_{UAV}, {\dot{\textbf{x}}_{UAV}}^{k-1})$ \\[+0.4em]
	    \end{tabular}
	}
	\KwOut{\\
	    \begin{tabular}{ll}
	        UAV state:      & \{${\textbf{x}^{k}_{UAV}}, \:\dot{\textbf{x}}^{k}_{UAV}\}$ \\[+0.2em]
	    \end{tabular}
	}  
	%
	\SetKwFunction{FSub}{$uav\_tracking\left(\mathcal{P}, \:I, \:{\textbf{x}^{k-1}_{UAV}}, \:\dot{\textbf{x}}^{k-1}_{UAV}\right)$}
    \SetKwProg{Fn}{Function}{:}{}
    \BlankLine
    \Fn{\FSub}
    {
            \vspace{.3em}
            \begin{tabular}{ll}
                UAV pos estimation:     & $\hat{\textbf{x}}_{\tiny UAV}^{k} = \mathcal{KF}^{}_{prediction}(\textbf{x}^{k-1}_{UAV});$ \\[+0.4em]
                Generate KD Tree:       & $kdtree \leftarrow \mathcal{P}^{'};$ \\[+0.4em]
                UAV points:             & $\mathcal{P}^{k}_{UAV} = KNN(kdtree, \: \hat{\textbf{x}}_{UAV}^{k});$ \\[+0.4em]
                UAV measurement:   & ${\textbf{z}^{k}_{UAV}} = \frac{1}{\lvert\mathcal{P}^{k}_{UAV}\rvert}\sum_{x\in\mathcal{P}^{k}_{UAV}} x;$\\[+0.4em]
                UAV state estimation:   & ${\textbf{x}^{k}_{UAV}} = \mathcal{KF}^{}_{update}(\textbf{z}^{k}_{UAV});$ 
                \\
            \end{tabular}
            \KwRet ${\textbf{x}^{k}_{UAV}};$
    }

    \While{new $\mathcal{P}_{k}\left(I\right)$}{
        ${\textbf{x}^{k}_{UAV}} =  uav\_tracking\left(\mathcal{P}_{k}(I), I, \textbf{x}^{k-1}_{}, \dot{\textbf{x}}^{k-1}_{}\right);$ \\
    }
	
\end{algorithm}

%% file: sections/04_Experiments.tex
\section{Experimental Results}

\begin{figure}
    \centering
    \includegraphics[width=\textwidth]{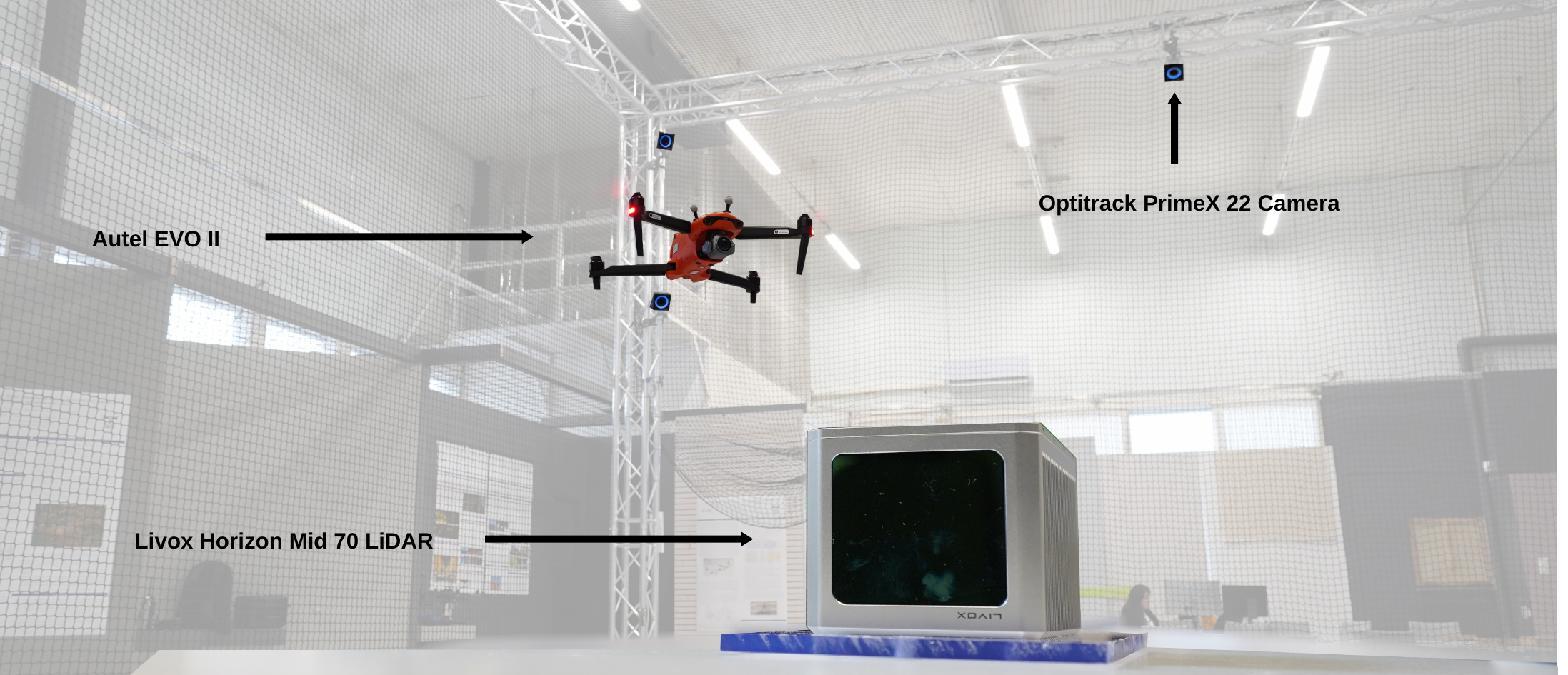}
    \caption{Experimental hardware setup.}
    \label{fig:setup}
\end{figure}

The experimental platform shown in Fig.~\ref{fig:setup} consists of a Livox Horizon LiDAR with a FoV of $81.7\degree\times25.1\degree$, which is able to output scanned point clouds up to 100\,Hz featuring a non-repetitive pattern. An external position system is used to validate the extracted trajectories.

\begin{figure}
    \centering
    \scriptsize{\input{tex/ape_boxplot.tex}}
    \caption{Absolute Position Error for different tracking methods.}
    \label{fig:ape_boxplot}
\end{figure}
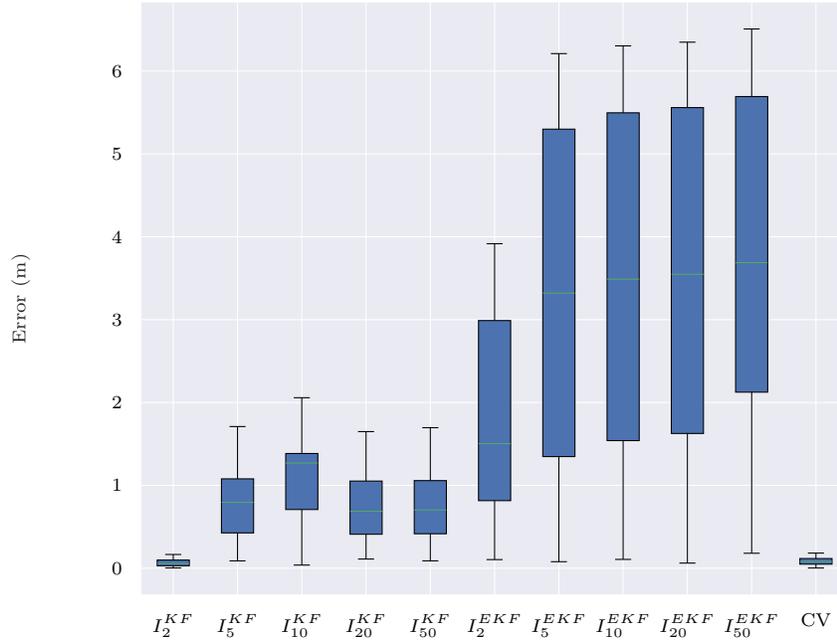

We tested the ability of the algorithm to track a UAV through a trajectory in a large open area, at distances ranging from 2\,m to over 20\,m from the LiDAR scanner, at variable speeds and directions. This testing allows us to evaluate the algorithm's performance in a realistic scenario. We performed a quantitative analysis of the Absolute Position Error (APE) based on the ground truth. The main results are summarized in Fig.~\ref{fig:ape_boxplot}, which shows the distribution of APE for different tracking modalities. To enable a fair comparison between trajectories, we transform all trajectories into the reference frame of the ground truth coordinates.

We also provide a comprehensive comparison of the APE for different UAV tracking modalities in terms of integration time ranges. Table~\ref{tab:ape_kf} shows the APE for tracking using both a Linear (KF) and an Extended Kalman Filter (EKF), as well as "tracking by detection" using only a CV model. The KF with a CV model performs better than the more complex EKF.

\input{tables/ape_table_kf}

\begin{figure}[H]
    \centering
    \includegraphics[width=\textwidth, height=\textheight]{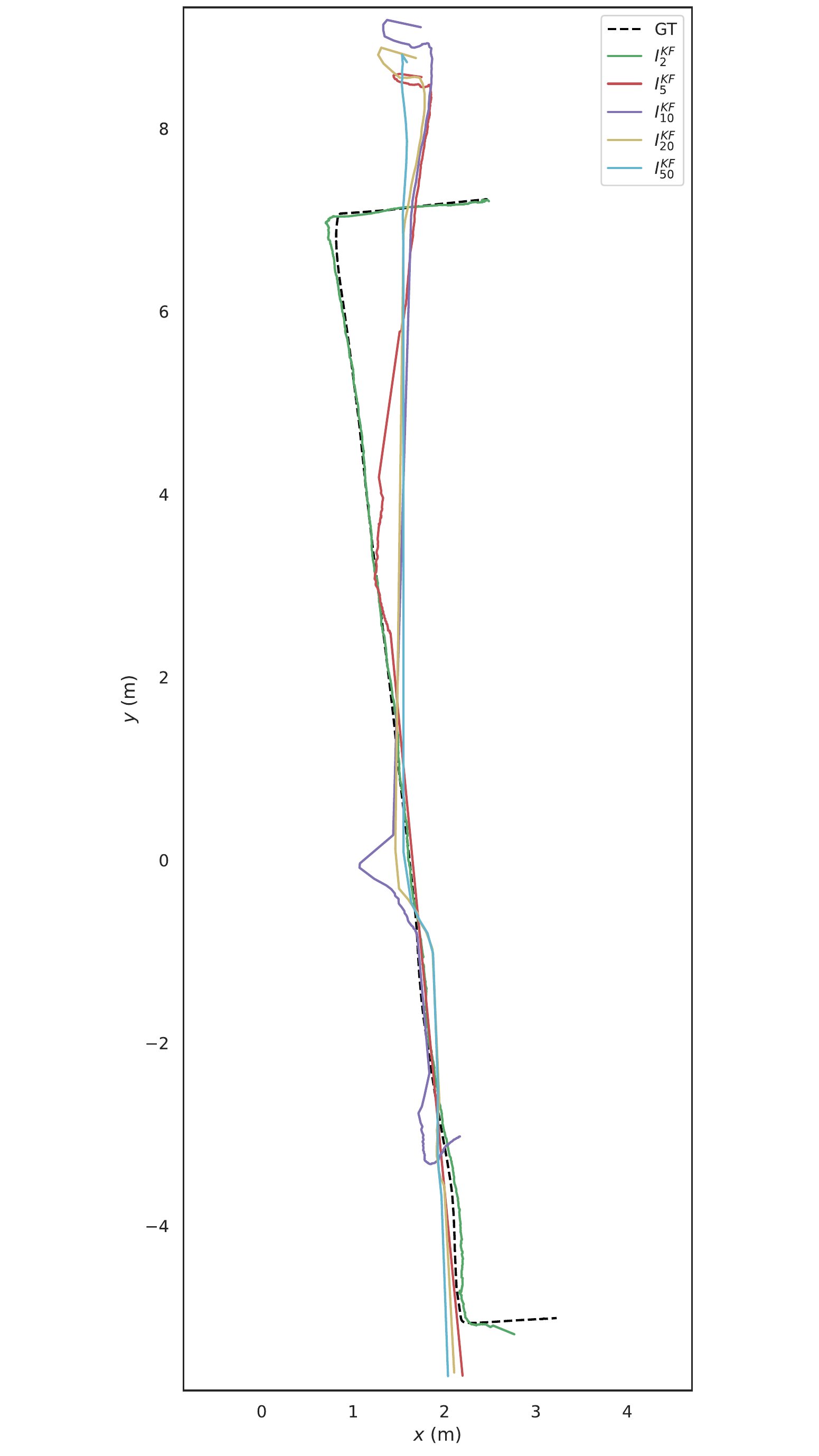}
    \caption{Comparison of the trajectories generated with the linear Kalman Filter (KF) against ground truth (GT) trajectories.}\label{fig:trajectories_single_integration_kf}
\end{figure}

\begin{figure}[H]
    \centering
    \includegraphics[width=0.8\textwidth, height=\textheight]{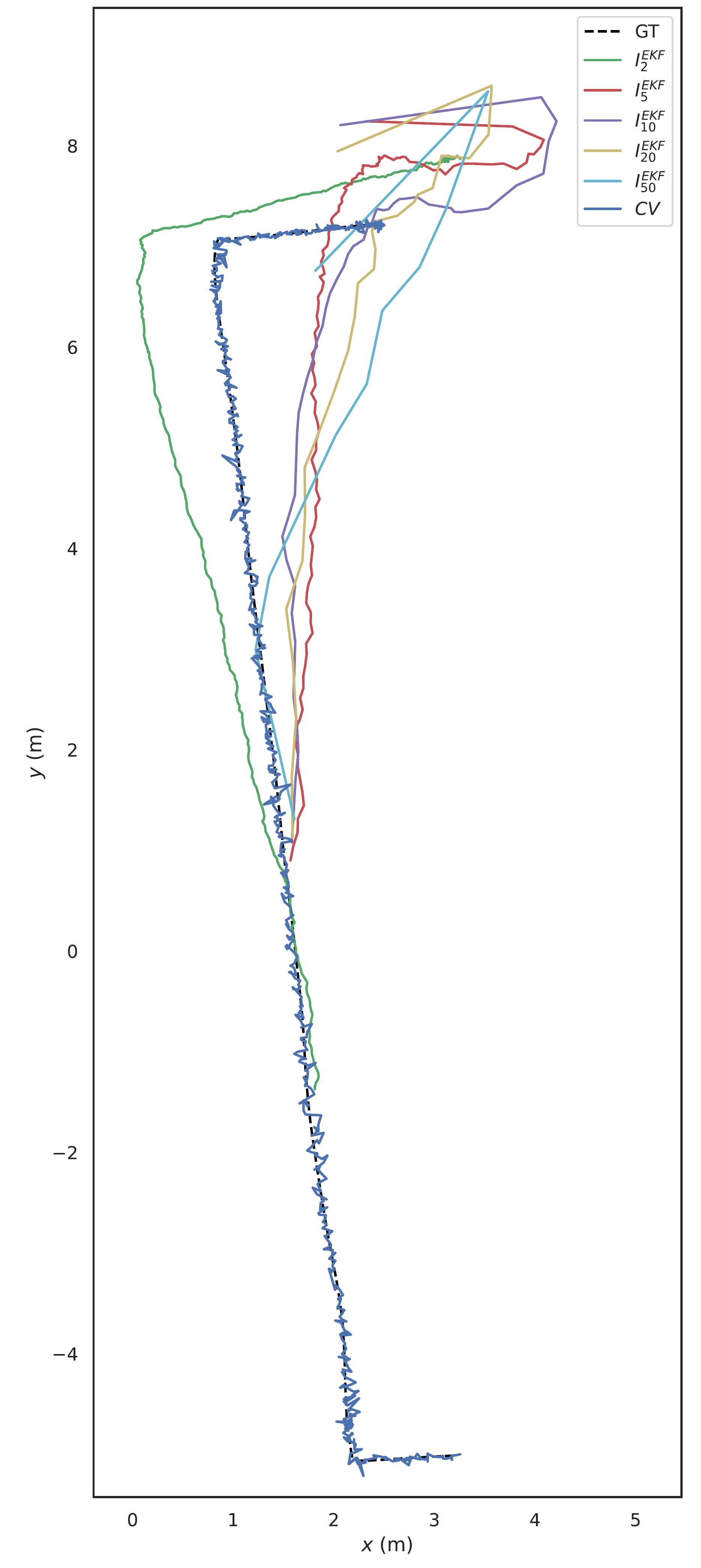}
    \caption{Comparison of the trajectories generated with the Extended Kalman Filter (EKF) and "tracking by detection" using the Constant Velocity (CV) model against ground truth (GT) trajectories.}\label{fig:trajectories_single_integration_ekf}
\end{figure}

To supplement the quantitative trajectory analysis, we provide a visualization of the trajectories obtained using different integration time ranges and methods. Figs.~\ref{fig:trajectories_single_integration_kf} and \ref{fig:trajectories_single_integration_ekf} show the trajectories for different integration time ranges and tracking modalities.

The majority of methods fail to reconstruct the overall trajectory, with only "tracking by detection" using the CV model being able to consistently track the target, although the trajectory itself is noisy. These results suggest that while "tracking by detection" using a CV model is the only method that consistently tracks the target throughout its trajectory, integrating 2 scans using a KF achieves lower position errors, making this a promising alternative for similar scenarios.

%% file: tex/ape_boxplot.tex
\begin{tikzpicture}

  \definecolor{darkslategray38}{RGB}{38,38,38}
  \definecolor{lightgray204}{RGB}{204,204,204}
  \definecolor{lavender234234242}{RGB}{234,234,242}
  \definecolor{mediumseagreen85168104}{RGB}{85,168,104}
  \definecolor{steelblue76114176}{RGB}{76,114,176}

  \begin{axis}[
    width=\figurewidth,
    axis background/.style={fill=lavender234234242},
    axis line style={white},
    tick align=outside,
    x grid style={white},
    xmajorgrids,
    xmajorticks=true,
    xmin=0.5, xmax=11.5,
    xtick style={draw=none},
    xtick={1,2,3,4,5,6,7,8,9,10,11},
    xticklabels={
      \(\displaystyle I_{2}^{KF}\),
      \(\displaystyle I_{5}^{KF}\),
      \(\displaystyle I_{10}^{KF}\),
      \(\displaystyle I_{20}^{KF}\),
      \(\displaystyle I_{50}^{KF}\),
      \(\displaystyle I_{2}^{EKF}\),
      \(\displaystyle I_{5}^{EKF}\),
      \(\displaystyle I_{10}^{EKF}\),
      \(\displaystyle I_{20}^{EKF}\),
      \(\displaystyle I_{50}^{EKF}\),
      CV
    },
    y grid style={white},
    ylabel=\textcolor{darkslategray38}{Error (m)},
    ylabel style={yshift=3.5ex},
    ymajorgrids,
    ymajorticks=true,
    ymin=-0.322926483622749, ymax=6.83425775468719,
    ytick style={draw=none}
    ]
    \path [draw=black, fill=steelblue76114176]
    (axis cs:0.75,0.0297829564134055)
    --(axis cs:1.25,0.0297829564134055)
    --(axis cs:1.25,0.0983892381625161)
    --(axis cs:0.75,0.0983892381625161)
    --(axis cs:0.75,0.0297829564134055)
    --cycle;
    \addplot [black]
    table {%
    1 0.0297829564134055
    1 0.00312699660695309
    };
    \addplot [black]
    table {%
    1 0.0983892381625161
    1 0.165531605790929
    };
    \addplot [black]
    table {%
    0.875 0.00312699660695309
    1.125 0.00312699660695309
    };
    \addplot [black]
    table {%
    0.875 0.165531605790929
    1.125 0.165531605790929
    };
    \path [draw=black, fill=steelblue76114176]
    (axis cs:1.75,0.426219921242237)
    --(axis cs:2.25,0.426219921242237)
    --(axis cs:2.25,1.07855259972683)
    --(axis cs:1.75,1.07855259972683)
    --(axis cs:1.75,0.426219921242237)
    --cycle;
    \addplot [black]
    table {%
    2 0.426219921242237
    2 0.0886013630683447
    };
    \addplot [black]
    table {%
    2 1.07855259972683
    2 1.70980815232771
    };
    \addplot [black]
    table {%
    1.875 0.0886013630683447
    2.125 0.0886013630683447
    };
    \addplot [black]
    table {%
    1.875 1.70980815232771
    2.125 1.70980815232771
    };
    \path [draw=black, fill=steelblue76114176]
    (axis cs:2.75,0.708544313138387)
    --(axis cs:3.25,0.708544313138387)
    --(axis cs:3.25,1.38442913206174)
    --(axis cs:2.75,1.38442913206174)
    --(axis cs:2.75,0.708544313138387)
    --cycle;
    \addplot [black]
    table {%
    3 0.708544313138387
    3 0.0377498002076827
    };
    \addplot [black]
    table {%
    3 1.38442913206174
    3 2.05798919120728
    };
    \addplot [black]
    table {%
    2.875 0.0377498002076827
    3.125 0.0377498002076827
    };
    \addplot [black]
    table {%
    2.875 2.05798919120728
    3.125 2.05798919120728
    };
    \path [draw=black, fill=steelblue76114176]
    (axis cs:3.75,0.410021046462895)
    --(axis cs:4.25,0.410021046462895)
    --(axis cs:4.25,1.05174892201314)
    --(axis cs:3.75,1.05174892201314)
    --(axis cs:3.75,0.410021046462895)
    --cycle;
    \addplot [black]
    table {%
    4 0.410021046462895
    4 0.110602501916276
    };
    \addplot [black]
    table {%
    4 1.05174892201314
    4 1.64873677419778
    };
    \addplot [black]
    table {%
    3.875 0.110602501916276
    4.125 0.110602501916276
    };
    \addplot [black]
    table {%
    3.875 1.64873677419778
    4.125 1.64873677419778
    };
    \path [draw=black, fill=steelblue76114176]
    (axis cs:4.75,0.415925034385366)
    --(axis cs:5.25,0.415925034385366)
    --(axis cs:5.25,1.0573012550807)
    --(axis cs:4.75,1.0573012550807)
    --(axis cs:4.75,0.415925034385366)
    --cycle;
    \addplot [black]
    table {%
    5 0.415925034385366
    5 0.0888383265474599
    };
    \addplot [black]
    table {%
    5 1.0573012550807
    5 1.69602601687129
    };
    \addplot [black]
    table {%
    4.875 0.0888383265474599
    5.125 0.0888383265474599
    };
    \addplot [black]
    table {%
    4.875 1.69602601687129
    5.125 1.69602601687129
    };
    \path [draw=black, fill=steelblue76114176]
    (axis cs:5.75,0.815580788757161)
    --(axis cs:6.25,0.815580788757161)
    --(axis cs:6.25,2.98941915671617)
    --(axis cs:5.75,2.98941915671617)
    --(axis cs:5.75,0.815580788757161)
    --cycle;
    \addplot [black]
    table {%
    6 0.815580788757161
    6 0.10384373386916
    };
    \addplot [black]
    table {%
    6 2.98941915671617
    6 3.9167645224361
    };
    \addplot [black]
    table {%
    5.875 0.10384373386916
    6.125 0.10384373386916
    };
    \addplot [black]
    table {%
    5.875 3.9167645224361
    6.125 3.9167645224361
    };
    \path [draw=black, fill=steelblue76114176]
    (axis cs:6.75,1.34716592364738)
    --(axis cs:7.25,1.34716592364738)
    --(axis cs:7.25,5.29959016184625)
    --(axis cs:6.75,5.29959016184625)
    --(axis cs:6.75,1.34716592364738)
    --cycle;
    \addplot [black]
    table {%
    7 1.34716592364738
    7 0.078446012094099
    };
    \addplot [black]
    table {%
    7 5.29959016184625
    7 6.21153242607861
    };
    \addplot [black]
    table {%
    6.875 0.078446012094099
    7.125 0.078446012094099
    };
    \addplot [black]
    table {%
    6.875 6.21153242607861
    7.125 6.21153242607861
    };
    \path [draw=black, fill=steelblue76114176]
    (axis cs:7.75,1.54000316886235)
    --(axis cs:8.25,1.54000316886235)
    --(axis cs:8.25,5.49676101833166)
    --(axis cs:7.75,5.49676101833166)
    --(axis cs:7.75,1.54000316886235)
    --cycle;
    \addplot [black]
    table {%
    8 1.54000316886235
    8 0.106390360384218
    };
    \addplot [black]
    table {%
    8 5.49676101833166
    8 6.30488254471562
    };
    \addplot [black]
    table {%
    7.875 0.106390360384218
    8.125 0.106390360384218
    };
    \addplot [black]
    table {%
    7.875 6.30488254471562
    8.125 6.30488254471562
    };
    \path [draw=black, fill=steelblue76114176]
    (axis cs:8.75,1.62632747271418)
    --(axis cs:9.25,1.62632747271418)
    --(axis cs:9.25,5.55980691840129)
    --(axis cs:8.75,5.55980691840129)
    --(axis cs:8.75,1.62632747271418)
    --cycle;
    \addplot [black]
    table {%
    9 1.62632747271418
    9 0.0616746312656346
    };
    \addplot [black]
    table {%
    9 5.55980691840129
    9 6.34960648845829
    };
    \addplot [black]
    table {%
    8.875 0.0616746312656346
    9.125 0.0616746312656346
    };
    \addplot [black]
    table {%
    8.875 6.34960648845829
    9.125 6.34960648845829
    };
    \path [draw=black, fill=steelblue76114176]
    (axis cs:9.75,2.1248437280742)
    --(axis cs:10.25,2.1248437280742)
    --(axis cs:10.25,5.69306661606805)
    --(axis cs:9.75,5.69306661606805)
    --(axis cs:9.75,2.1248437280742)
    --cycle;
    \addplot [black]
    table {%
    10 2.1248437280742
    10 0.180548970667889
    };
    \addplot [black]
    table {%
    10 5.69306661606805
    10 6.50893119840037
    };
    \addplot [black]
    table {%
    9.875 0.180548970667889
    10.125 0.180548970667889
    };
    \addplot [black]
    table {%
    9.875 6.50893119840037
    10.125 6.50893119840037
    };
    \path [draw=black, fill=steelblue76114176]
    (axis cs:10.75,0.0488397148765506)
    --(axis cs:11.25,0.0488397148765506)
    --(axis cs:11.25,0.116689110440969)
    --(axis cs:10.75,0.116689110440969)
    --(axis cs:10.75,0.0488397148765506)
    --cycle;
    \addplot [black]
    table {%
    11 0.0488397148765506
    11 0.00240007266406687
    };
    \addplot [black]
    table {%
    11 0.116689110440969
    11 0.182197424653785
    };
    \addplot [black]
    table {%
    10.875 0.00240007266406687
    11.125 0.00240007266406687
    };
    \addplot [black]
    table {%
    10.875 0.182197424653785
    11.125 0.182197424653785
    };
    \addplot [mediumseagreen85168104]
    table {%
    0.75 0.0567228727248979
    1.25 0.0567228727248979
    };
    \addplot [mediumseagreen85168104]
    table {%
    1.75 0.7947337010503
    2.25 0.7947337010503
    };
    \addplot [mediumseagreen85168104]
    table {%
    2.75 1.27001209771321
    3.25 1.27001209771321
    };
    \addplot [mediumseagreen85168104]
    table {%
    3.75 0.687452739891269
    4.25 0.687452739891269
    };
    \addplot [mediumseagreen85168104]
    table {%
    4.75 0.703120698238754
    5.25 0.703120698238754
    };
    \addplot [mediumseagreen85168104]
    table {%
    5.75 1.50356596394448
    6.25 1.50356596394448
    };
    \addplot [mediumseagreen85168104]
    table {%
    6.75 3.3203797558113
    7.25 3.3203797558113
    };
    \addplot [mediumseagreen85168104]
    table {%
    7.75 3.48825415965513
    8.25 3.48825415965513
    };
    \addplot [mediumseagreen85168104]
    table {%
    8.75 3.54780859471506
    9.25 3.54780859471506
    };
    \addplot [mediumseagreen85168104]
    table {%
    9.75 3.68811426251003
    10.25 3.68811426251003
    };
    \addplot [mediumseagreen85168104]
    table {%
    10.75 0.0791649911112496
    11.25 0.0791649911112496
    };
  \end{axis}

\end{tikzpicture}

%% file: tables/ape_table_kf.tex
\begin{table}[H]
    \centering
    \caption{Absolute Position Error (APE) for different tracking modalities: linear Kalman Filter (KF), Extended Kalman Filter (EKF) and "tracking by detection" with Constant Velocity (CV) model.}
    \resizebox{\textwidth}{!}{
        \begin{tabular}{@{}lccccccccccc@{}}
            \toprule
            \textbf{Tracking Method} & $I_{2}^{KF}$ & $I_{5}^{KF}$ & $I_{10}^{KF}$ & $I_{20}^{KF}$ & $I_{50}^{KF}$ & $I_{2}^{EKF}$ & $I_{5}^{EKF}$ & $I_{10}^{EKF}$ & $I_{20}^{EKF}$ & $I_{50}^{EKF}$ & CV\\
            \midrule
            \textbf{APE (m)} & 0.0567 & 0.7947 & 1.270 & 0.6874 & 0.7031 & 1.5035 & 3.3204 & 3.4882 & 3.5478 & 3.6881 & 0.0791\\ \bottomrule
        \end{tabular}
    }
    \label{tab:ape_kf}
\end{table}

%% file: sections/05_Conclusion.tex
\section{Conclusion and Future Work}

In this study, we investigated the performance of different tracking modalities for UAVs using a solid-state LiDAR with non-repetitive scan patterns. Our results indicate that integrating multiple scans (i.e., scanning at variable frequencies) using a Kalman Filter can effectively reduce the absolute position error with respect to only using a Constant Velocity model, making it a viable option for tracking UAVs in challenging environments. The Constant Velocity model combined with a linear Kalman Filter achieves better results than the Extended Kalman Filter, despite the latter's added complexity. However, "tracking by detection" using a Constant Velocity model was the only method that consistently tracked the UAV throughout its trajectory.

In future research, we plan to investigate the potential of dynamically adjusting and integrating LiDAR scans to enhance the robustness and consistency of the UAV tracking algorithm. Specifically, we aim to explore adaptive scan patterns, optimized frame integration times, and other advanced techniques to improve the accuracy and reliability of the tracking system in challenging environments.